\documentclass{sn-jnl}

\usepackage{graphicx}%
\usepackage{multirow}%
\usepackage{amsmath,amssymb,amsfonts}%
\usepackage{amsthm}%
\usepackage{mathrsfs}%
\usepackage[title]{appendix}%
\usepackage{textcomp}%
\usepackage{manyfoot}%
\usepackage{booktabs}%
\usepackage{algorithm}%
\usepackage{algorithmicx}%
\usepackage{algpseudocode}%
\usepackage{listings}%
\usepackage[numbers]{natbib}
\usepackage[table]{xcolor}
\usepackage{cleveref} 
\UseRawInputEncoding

\graphicspath{ {Figures/} }
\begin{document}

\title[Article Title]{ConstellationNet: Reinventing Spatial Clustering through GNNs}


\author*[1]{\fnm{Aidan} \sur{Gao}}\email{aidangao345@gmail.com}

\author[2]{\fnm{Junhong} \sur{Lin}}\email{junhong@mit.edu}
\equalcont{Mentored this work.}

\affil*[1]{ \orgname{MIT PRIMES}, \orgaddress{\street{32 Vassar Street}, \city{Cambridge}, \postcode{02139}, \state{MA}, \country{USA}}}

\affil[2]{\orgname{MIT},  \orgaddress{\street{32 Vassar Street}, \city{Cambridge}, \postcode{02139}, \state{MA}, \country{USA}}}


\abstract{Spatial clustering is a crucial field, finding universal use across criminology, pathology, and urban planning. However, most spatial clustering algorithms cannot pull information from nearby nodes and suffer performance drops when dealing with higher dimensionality and large datasets, making them suboptimal for large-scale and high-dimensional clustering. Due to modern data growing in size and dimension, clustering algorithms become weaker when addressing multifaceted issues. To improve upon this, we develop ConstellationNet, a convolution neural network(CNN)-graph neural network(GNN) framework that leverages the embedding power of a CNN, the neighbor aggregation of a GNN, and a neural network's ability to deal with batched data to improve spatial clustering and classification with graph augmented predictions. ConstellationNet achieves state-of-the-art performance on both supervised classification and unsupervised clustering across several datasets, outperforming state-of-the-art classification and clustering while reducing model size and training time by up to tenfold and improving baselines by 10 times. Because of its fast training and powerful nature, ConstellationNet holds promise in fields like epidemiology and medical imaging, able to quickly train on new data to develop robust responses. }

\keywords{Clustering, Graph Neural Networks, Convolutional Neural Networks, Spatial Clustering, Clustering, Unsupervised Training}



\maketitle

\section{Introduction}\label{sec1}

Clustering, the grouping of data points based on similarity, has remained a crucial topic across many fields, finding applications in search engines\cite{search, search2}, medical imaging\cite{cancer, medical2}, and anomaly detection\cite{anomaly1, anomaly2}. Although clustering can be performed on almost any dataset, spatial clustering, which focuses on clustering data points in space, acts as the primary and most crucial clustering method because most datatypes, images, or three-dimensional molecular structures can be vectorized into spatial contexts. Traditional spatial clustering methods, like K-means or dbscan, have been widely used for processing and classifying large-scale spatial data, and recent advancements have improved density-based clustering while introducing spectral and covariance-based clustering\cite{kmeans, dbscan}. While these methods have improved clustering robustness significantly, most methods still often face three significant challenges when dealing with high-dimensional data and large datasets:
\begin{itemize}
  \item[(1)] \textbf{The Curse of Dimensionality.} Since data becomes sparser in higher dimensions, models have a harder time clustering these points based on distance.\cite{curse1, curse2}
  \item[(2)] \textbf{Computational Inefficiency.} Due to most models having around $O(n^2)$ time complexity and being superpolynomial in worse-case scenarios, larger datasets become much harder for clustering methods to handle.\cite{time}
  \item[(3)]\textbf{Effetive Use of Neighborhood Information.} Since most spatial clustering models are based on the distance between points alone, they do not consider points' relations with each other, especially in dense groups, which can weaken their power when dealing with irregular data. 

\end{itemize}

In a different realm, with the recent rise in machine learning, many different types of neural networks have been developed on various mediums, like images or languages, with some work being done into unsupervised classification, which can be analogized to clustering\cite{cnn1,cnn2}. However, due to only operating on the features of a single data point, these don't act similarly to clustering and suffer from the same limited scope. Graph Neural Networks (GNNs) have emerged as a powerful tool for learning from graph-structured data by leveraging information from both nodes and edges\cite{introduction}, showing promising results in various applications such as social network analysis\cite{socialnetwork}, drug discovery\cite{drug}, and human object interaction\cite{human}. Due to their ability to aggregate information from neighbors and edges, GNNs have been applied to graph clustering tasks and have the potential to be used for spatial clustering tasks. However, Their primary strength of extrapolating information from edge and neighbor data restricts their application to graph-structured data, constrained primarily by the absence of edges between nodes in spatial data.

To address these challenges, this paper introduces ConstellationNet, a novel CNN-GNN spatial clustering framework designed to effectively and quickly cluster high-dimensional and large-scale spatial data. By integrating Convolutional Neural Networks (CNNs) with Graph Neural Networks (GNNs), ConstellationNet leverages the local feature extraction of CNNs and the neighborhood aggregation of GNNs, allowing the model to effectively extrapolate clusters of arbitrary shape and density. By constructing a weighted K-Nearest Neighbors (KNN) graph from spatial data, the model creates graph data from spatial data, using the GNN's unique aggregation to counteract the curse of dimensionality on created edges. To better distinguish features, ConstellationNet introduces an innovative CNN-GNN data-passing mechanism, passing the CNN's output to the GNN, which gives the GNN more distinguishable features to cluster on, improving performance and power.
Additionally, ConstellationNet incorporates the Deep Modularity Network (DMoN) operator and minibatching as further enhancements\cite{dmon}\cite{neighborloader}. The DMoN operator provides an unsupervised pooling operator that optimizes cluster assignments and a loss, facilitating end-to-end clustering without requiring labels and mimicking traditional spatial clustering behavior. Furthermore, by processing subsets of the graph, ConstellationNet maintains high clustering accuracy while enabling scalability to larger datasets via more computational efficiency. In both supervised and unsupervised contexts, ConstellationNet achieves superior clustering performance compared to state-of-the-art methods, improving baselines by up to 10 times while beating state-of-the-art with up to tenfold reductions in parameters and training time. 

This paper's contributions are summarized as follows:
\begin{itemize}
    \item \textbf{Algorithm:} We propose ConstellationNet, a new framework that integrates CNNs and GNNs to perform spatial clustering on high-dimensional, large-scale datasets. We demonstrate the methodology for combining edge construction techniques with GNNs, enabling the application of graph-based learning methods to non-graph spatial data.
    \item \textbf{Extension:} We extend several graph-based and image-based methods to a spatial context, proving their viability and potential for future use in the area. 
    \item \textbf{Evaluation:} We perform extensive experiments across several datasets and ablation studies to demonstrate the performance of each part of the framework, achieving state-of-the-art results in both supervised and unsupervised contexts with lower investments in memory and time. 
\end{itemize}

\section{Background}\label{sec2}

\subsection{Spatial Clustering}

Spatial clustering aims to group spatial data points based on location and distance, uncovering structures within spatial datasets. Traditional methods have been broadly categorized into centroid-based, density-based, and distribution-based approaches \cite{han2011data}. Centroid-based methods, such as K-means \cite{lloyd1982least}, partition data into a predefined number of clusters by minimizing the distance between data points and cluster centroids. Density-based methods like DBSCAN \cite{ester1996density} identify clusters as areas of high point density. Distribution-based methods assume data are generated from a mixture of underlying probability distributions and clusters as such\cite{reynolds2009gaussian}.

Recent research in spatial clustering has focused on enhancing traditional methods and developing new algorithms. New algorithms include spectral Clustering \cite{ng2002spectral}, which uses eigenvectors of a similarity matrix to perform dimensionality reduction before clustering, and Deep learning approaches like Deep Embedded Clustering (DEC) \cite{xie2016unsupervised} which learns feature representations and cluster assignments.
Improved baselines include OPTICS\cite{ankerst1999optics} and ST-DBSCAN \cite{birant2007st},  density-based methods that address the limitation of DBSCAN in detecting clusters with varying density. However, these new methods still ultimately suffer from the same curse of dimensionality due to the inherent lack of separation that high dimensional data creates, leading to worse performances on datasets with higher dimensionality, like STL and CIFAR.

\subsection{Graph Neural Networks}

\begin{wrapfigure}{r}{8.1cm}
\begin{center}
\includegraphics[width=8cm]{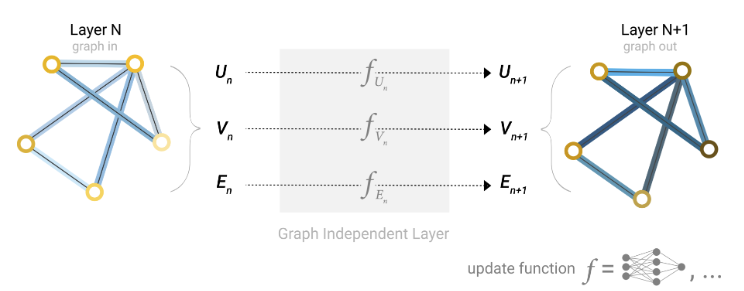}
\caption{A visualization of a GNN transform function\cite{introduction}.}
\label{fig1}
\end{center}
\end{wrapfigure}

Graph neural networks (GNNs) are a class of neural networks specifically designed to perform inference on data described by graphs. Unlike traditional neural networks, which use a series of transforms on the information expressed through a vector, GNNs instead process similar vector data that describes graph information, using neural networks to transform their features, as in \cref{fig1}. 

This transformation happens to node features, edge features, and graph information, acting similarly to other neural networks in highlighting features and predicting properties. However, due to the connected nature of a graph, GNNs are much more versatile and able to aggregate information from different parts to update representations. This can be done in predictions, where message passing allows edge embeddings to predict node classifications and vice versa, or in training, where each node's representation is updated by combining its features with the aggregated information from its neighbors, allowing the network to learn localized patterns within the graph \cite{kipf2017semi}.

Besides the basic GNN, several GNN architectures have been proposed, including Graph Convolutional Networks (GCNs) \cite{kipf2017semi}, Graph Attention Networks (GATs) \cite{velivckovic2018graph}, and Graph Isometric Networks \cite{GIN}. These models differ primarily in how they aggregate information from neighboring nodes. GCNs use a localized approach to aggregate information with convolutional operations based on a convolution neural network, GATs introduce attention mechanisms that factor in edge weights to assign weights to neighbors and create selected aggregations, and GINs use a Weisfeiler-Lehman graph isomorphism test and a separate multilayer perception to aggregate data. 

In spatial clustering, GNNs have had somewhat limited applications to the topic. Currently, most research into applying GNNs to spatial clustering has been focused on cell transcriptomics, with most methods establishing the construction of a graph through distance-based edge generation. We highlight two papers discussing the topic of applying GNNs toward spatial clustering: Learning Hierarchical Graph Neural Networks for Image Clustering and Cell Clustering for Spatial Transcriptomics Data with Graph Neural Networks\cite{ex1}\cite{ex2}. Graph Neural Networks for Image Clustering establishes several precedents for this paper, mainly using K-Nearest Neighbors for graph creation. 

\section{Methods}\label{sec3}
This section describes the Preliminaries and architecture of ConstellationNet. The main libraries used are PyTorch and PyTorch Geometric. 
\subsection{Preliminaries}

\subsubsection{Dataset Construction}
\label{dataset}
\begin{figure}[h]
\begin{center}
\includegraphics[width=13cm]{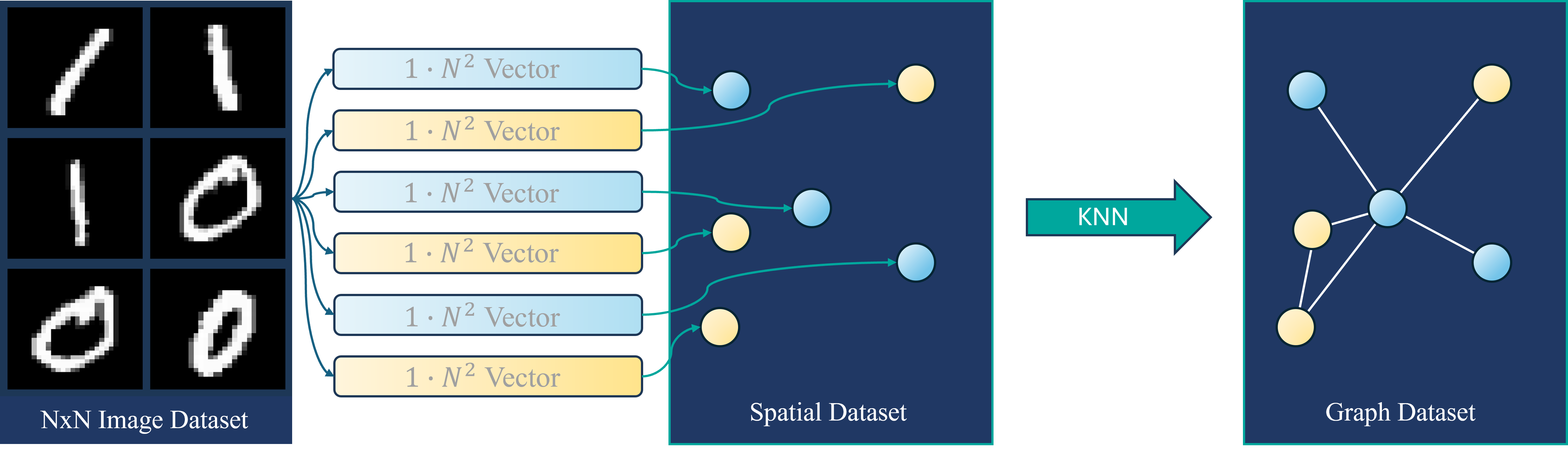}
\caption{The dataset transformation pipeline visualized on six samples from the MNIST dataset, moving from images to nodes on a graph\cite{mnist}.}
\label{fig2}
\end{center}
\end{figure}

To transform a spatial dataset into a graph for clustering, all images from the dataset are extracted and turned into one-dimensional vectors of values based on the process used in PECANN\cite{pecann}. For instance, a single image in MNIST is transformed into a shape $(1,728)$ vector with the 728 values corresponding to the pixel values concatenated. Each data point is then treated as a node, and the entire dataset is treated as a graph, allowing the node feature array to be created by stacking all images. Once a node feature array is obtained, a K-nearest neighbors algorithm is run on the data with a varying number of neighbors, denoted as the $K$ value and an edge index is created based on the indexes returned from the K-nearest neighbors. An additional edge weight array is then created by inverting the distance between each edge created by the K-nearest neighbors. The data transformation process is seen in \cref{fig2}.

\subsubsection{Deep Modularity Network operator}
To define a loss that ConstellationNet can train on during unsupervised clustering, the Deep Modularity Network (DMON) operator introduced by Google Research is extended to a spatial application, defined as the following\cite{dmon}: $$\mathcal{L}_{\text{DMon}}(\mathbf{C}; \mathbf{A}) = -\frac{1}{2m} \operatorname{Tr}(\mathbf{C}^\top \mathbf{B} \mathbf{C}) + \frac{\sqrt{k}}{n} \left\| \sum_i \mathbf{C}_i^\top \right\|_F - 1$$Functioning as a pooling mechanism, the DMON operator integrates a differentiable clustering module that optimizes clustering assignments within the GNN framework through spectral modularity maximization, which seeks to maximize modularity based off of edge distribution differences between clusters and random distributions. The DMoN operator uses collapse regularization, a relaxed constraint on the soft clustering assignments, to prevent trivial assignments without compromising the clustering objective. 

Besides acting as a clustering and pooling function, the DMoN operator provides the main unsupervised losses that guide the framework during unsupervised training, being the spectral loss and collapse regularization loss that the operator returns. Besides these losses, ConstellationNet employs two additional auxiliary losses to improve model performance: orthogonality loss and clustering loss. Orthogonality loss maximizes orthogonality between rows and columns in weight matrices, improving model uniqueness, while clustering loss attempts to balance cluster sizes. 

\subsubsection{CNN Embedding}
Enabling GNNs and baselines to better extrapolate features and grouping clusters closer together, CNN embedding functions as a feature extractor through magnifying certain spatial features\cite{lecun1998gradient}. The extracted features are then pooled and projected into a lower-dimensional space via a fully connected layer, resulting in embeddings that serve as more distinct node features in the graph. 

CNN embedding can be done in two ways: training a CNN to classify and then removing its fully connected layer to create an embedding model, or specifically training an embedding model via projection losses like triple loss\cite{triple}. A benefit of CNN embedding is the fact that it can be done in both supervised and unsupervised manners, with triple loss frameworks being an example of supervised frameworks and DINO and Simclr being unsupervised frameworks. \cite{dino, simclr}

\subsection{ConstellationNet}
To address the issues with large-scale image clustering, We present a novel framework, ConstellationNet, that can be used both as a spatial embedding technique and as a standalone end-to-end clustering pipeline. ConstellationNet consists of two similar frameworks: a dynamic framework for classification and a static framework for clustering. 
\pagebreak
\begin{figure}[H]
\begin{center}
\includegraphics[width=13cm]{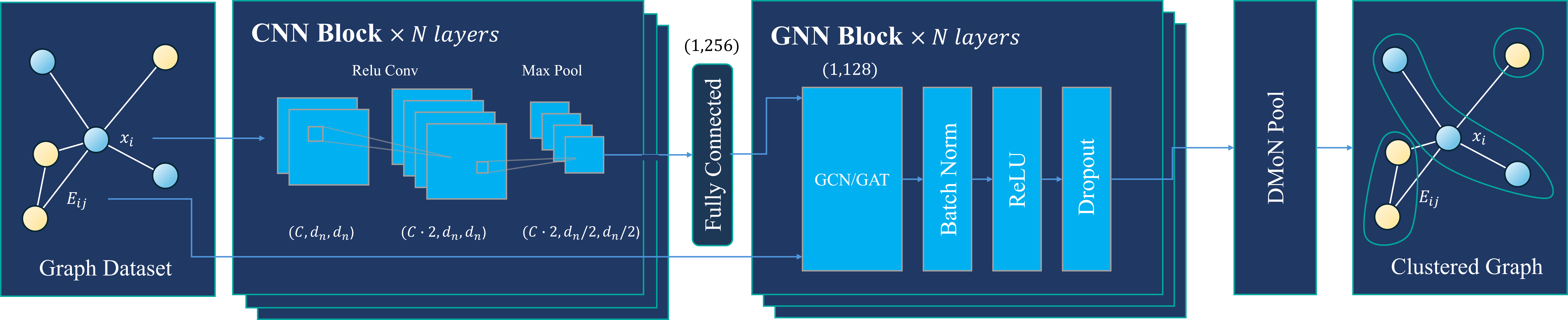}
\caption{The Supervised CNN-GNN pipeline, acting on the end product of dataset construction. $x_i$ Represents a node’s attribute, and $E_{ij}$ denotes an edge between node $x_i$ and $x_j$. $C$ represents the number of channels an image has, and $D_n$ represents the size of the image.}
\label{fig3}

\end{center}
\end{figure}

Supervised ConstellationNet is illustrated in \cref{fig3} above, consisting of a dynamic CNN-GNN framework with a final DMoN pool. Given a constructed graph dataset with spatial features $X = (x_1, x_2, x_3, \cdots, x_n) \in \mathbb{R}^{n \times d_n^2}$, ConstellationNet iteratively samples large neighborhoods to train on given a random cluster of nodes $X_{rand} \in \mathbb{R}^{b\times d_n^2}$ where $b$ is the batch size, resulting a final feature array $X_{rand} \in \mathbb{R}^{b\cdot d\cdot b_n\times d_n^2}$ where $b_n$ is the neighborhood size and $d$ is the neighborhood hop depth. This feature array is then passed through the CNN embedder that reduces dimensionality down to variable size, transforming the subsampled feature array into $X_{CNN} = (x_1^2 x_2^2, x_3^2, \cdots, x_n^2) \in \mathbb{R}^{b\cdot d \cdot b_n \times G}$ where $G$ is the final dimension of the CNN and the starting dimension of the GNN. This feature array and the subsampled portion of the edge index $E \in \mathbb{R}^{2 \times K\cdot n}$ are then passed into the GNN, which aggregates information before the DMoN pool finally clusters. We choose to use a residual edge index connection instead of building the edge index after reducing dimensionality with the CNN to maintain more information, allow the GNN to better aggregate distilled features based on original connections, and allow for dynamic updates. The main difference of this framework is its dynamic nature, functioning as one collective model. Because this framework operates in a supervised manner, both the CNN and GNN can be trained on supervised loss, allowing for abstract representations of data that can enhance the functionality of overall model more than separately training. To counteract the lack of connections between test data, we introduce graph-augmented classification, allowing ConstellationNet to take advantage of its training to augment predictions further. By connecting a test data point to the created train graph during prediction, Constellation establishes links to known data, building upon the training process and utilizing its data like never before. 

\begin{figure}[H]
\begin{center}
\includegraphics[width=13cm]{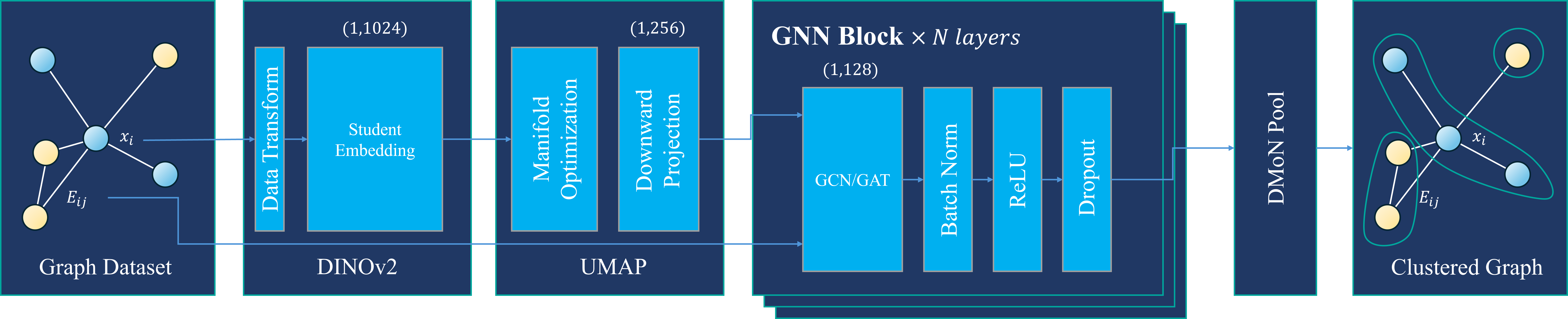}
\caption{The Unsupervised CNN-GNN pipeline, acting on the end product of dataset construction. Data is passed through two transforms before the GNN, which are static in this case.\protect\footnotemark}
\label{fig4}

\end{center}
\end{figure} 

The unsupervised ConstellationNet, seen in \cref{fig4} above, acts similarly to the supervised ConstellationNet, with a few changes. The same neighborhood sampling and residual edge connections are kept, leading the same arrays $X_{rand} \in \mathbb{R}^{b\cdot db_n\times d_n^2}$ and $E \in \mathbb{R}^{2 \times K\cdot n}$. However, unsupervised ConstellationNet is not a singular model but a pipeline through which data is passed because the DMoN operator and its losses are graph-focused, meaning that combined unsupervised training doesn't yield good results. Since an untrained CNN isn't good at feature separation to enhance clustering quality, spatial features are instead passed without neighborhoods through a large DINO v2 model, an unsupervised model pretrained on imagenet via a student and teacher framework, to transfer embed data into shape $X_{DINO} \in \mathbb{R}^{n\times 1024}$ \cite{dino}. To further improve the separation of clusters, Uniform Manifold Approximation and Projection(UMAP) is run through the embedded features to produce $X_{UMAP} \in \mathbb{R}^{n\times G}$. While empirically like other clustering algorithms, UMAP has demonstrated the ability to further enhance clusters once DINOv2 embeds features. Finally, after two embeddings, data is passed into the GNN and clustered via DMoN. 

One notable feature of both supervised and unsupervised ConstellationNet is the interchangeability of all parts and the ability for the entire framework to be used as a transform. Since the end classifier is the GNN with a DMoN cluster optimizer, by removing the DMoN clustering assignment, the entire framework functions as a data embedding, reducing dimensionality down to the variable hidden dimension of the GNN and creating a new dataset that any spatial clustering baseline can quickly cluster. Additionally, the CNN and UMAP embeddings can be used as a separate standalone transform, which allows the final GNN to be switched out for any spatial clustering method. 
\section{Experimentation}\label{sec4}
\subsection{Experimental Setup}

\subsubsection{Dataset}

We primarily conduct experiments on well-known image datasets, as images are a widely used form of spatial data. In a supervised setting, models are tested on the MNIST, CIFAR-10, and Imagnet image datasets, with MNIST being a black-and-white collection of handwritten digits while CIFAR-10 and Imagenet are both colored datasets of various objects\cite{mnist,cifar,imagenet}. In an unsupervised setting, models are tested on MNIST, CIFAR-10, CIFAR-100, and Oxford flowers\cite{mnist,cifar,flowers}. Due to Imagenet's large size, testing is somewhat inconclusive in a supervised setting, using only 100 thousand data points, and thus, the dataset is not used in unsupervised testing.

\subsubsection{Models}

Several GNNs are used, with the main models mentioned in the background being the GCN, GAT, and GIN, as well as the SuperGAT, a version of the GAT with self-supervised attention. All models are built on PyTorch Geometric, and most models use existing convolutions within the library, except for the NAGphormer, a graph transformer that uses the source code provided by the paper\cite{nagformer}. For the GCN, GAT, and GIN, each model is built through stacking a variable number of the layers consisting of the following: the chosen convolution followed by a batch norm, a Relu activation function, and a dropout layer of 30\%. The multilayer perception used for the GIN followed the MLP used in the paper, comprising two linear layers with a batch norm and Relu function between. During experiments, the two main properties that varied were the number of layers and the hidden dimension of the convolution and batch norms. After the final layer that transforms the data to the output dimension, the edge index is converted into a dense adjacency matrix and passed into the DMoN operator for clustering. 

\subsubsection{Metrics}
To evaluate models, we use accuracy, NMI, and ARI throughout unsupervised testing to match with standard metrics. We only used top-1 accuracy during supervised training to match state-of-the-art image classification. Note the accuracy used during supervised and unsupervised training are different types of accuracy, with the supervised being a strict top 1 accuracy while the unsupervised uses a clustering accuracy determined via a contingency matrix of checking if clusters with the same class are in the same cluster and if clusters of different classes are in different clusters. 

\subsubsection{Supervised Constants}
Training on all three vision datasets involved creating the graph as described in \cref{dataset} with a $K$ of 50 and minibatching through Pytorch Geometric's neighbor-loader, sampling 30 neighbors over two hops during training and all neighbors during testing with a starting batch size of 64. For the imagenet dataset, a ConvNet embeds data into 1024 dimensions before feeding into the model\cite{convnet}. Two different CNN architectures were used, one for the black and white images from MNIST and one for the RGB images from CIFAR 10. Both architectures expand from the input channel to 32, 64, and 128 channels over three convolutions, followed by a Relu activation and 2d maxpool on the later two convolutions. The MNIST CNN embeds the output down to 256, while the RGB model first embeds to 512 before embedding down to 256. The GNN model consists of the standard GNN architecture with a GCN convolution and a hidden size of 128 with five layers. Models were trained a variable number of epochs on the training portion of the dataset with negative log-likelihood, optimized by an Adam optimizer with a 0.001 learning rate and reduce learning rate on plateau scheduler with a patience of 5 epochs and a multiplication factor of 0.75 and tested using metrics described above.

\subsubsection{Unsupervised Constants}
Unsupervised training involves the same graph creation process and $K$ value of 50, with the same minibatching happening as in the supervised constants. A DINO v2 model is used to universally embed down to 1024 dimensions, and the UMAP then further embeds down to either 512 or 256 dimensions, which is the starting point of the GNN. Either a GAT or GCN is used as the GNN, and the hidden dimension is half of the starting dimension. Models were trained for up to 100 epochs on the entire dataset with DMoN losses and collapse regularization with 0.1 multiplication factor, optimized by an Adam optimizer with a 0.001 learning rate and reduce learning rate on plateau scheduler with a patience of 5 epochs and a multiplication factor of 0.75 and tested using metrics described above.

\subsubsection{Baselines}
\begin{wraptable}{l}{10.5cm}
\begin{tabular}{ |p{5cm}||p{2cm}|p{2cm}|}
\hline
Model &Accuracy &NMI \\
\hline
K-Means 10 cluster &0.906 &0.492 \\
DBScan 3eps 2 samples &0.836 &0.256 \\
Optic 2 samples 1k &0.854 &0.226  \\
Birch 10 clusters 10k &\cellcolor{black!25}0.925 &0.685 \\
Ward 10 clusters &0.914 &0.693  \\
Agglomerative 10 clusters &0.915 &\cellcolor{black!25}0.693 \\
HDBscan &0.839 &0.277  \\
\bottomrule

\end{tabular}
\caption{Baseline Accuracies and NMI for MNIST.}
\label{tab1}
\end{wraptable}

A collection of spatial clustering processes and results from current state-of-the-art clustering are collected to provide a point of comparison between models and traditional clustering processes. Baselines use most clustering methods from sci-kit learn's clustering page, as they comprise the most popular and powerful spatial clustering methods\cite{scikit}. These models include K-means, DBScan and HDBScan, OPTICS, ward and agglomerative clustering, and BIRCH. Out of these, Kmeans and Aggolmerative are typically used when comparing against GNNs, Kmeans due to its simplicity and Aggolmerative due to its being the best-performing baseline tested, as seen in \cref{tab1}. Apart from baselines, state-of-the-art is taken from the top performing model on papers with code pages for each of the datasets used, with supervised state-of-the-art coming from the image classification leaderboards, while the unsupervised state-of-the-art comes from image clustering leaderboards.

\subsection{Supervised Results}

\begin{table}[h]
    \centering
    \begin{tabular}{|l|l||l|l||l|l|}
    \hline
        MNIST: & ~ & CIFAR 10: & ~ & Imagenet: & ~ \\ \hline
        Model & Accuracy & Model & Accuracy & Model & Accuracy \\ \hline
        ConstellationNet & 99.96\% & ConstellationNet & 99.67\% & ConstellationNet* & 91.10\% \\ 
        Merging CNN & 99.87\% & Efficient Adaptive Ensemble & 99.61\% & OmniVec & 92.40\% \\ 
        EnsNet & 99.84\% & ViT-H/14 & 99.50\% & CoCa & 91\% \\ 
        Efficient-CapsNet & 99.84\% & DINOv2 & 99.50\% & Model Soups & 90.98\% \\ 
        SOPCNN & 99.83\% & µ2Net & 99.49\% & ViT-e & 90.90\% \\ \hline
    \end{tabular}
    \caption{ConstellationNet compared against state of the art on MNIST, CIFAR-10, and Imagenet. }
\label{tab2}
\end{table}
 Table \ref{tab2} lists the results of ConstellationNet across MNIST, CIFAR-10, and ImageNet, compared against image classification state-of-the-art on each dataset. Seen across all datasets, ConstellationNet demonstrates leading performance as indicated by its top rankings amongst state-of-the-art. On MNIST, ConstellationNet achieves an accuracy of 99.96\%, a near-perfect accuracy of 0.09\% better than the next state-of-the-art. Similarly, on CIFAR-10, ConstellationNet attains an accuracy of 99.67\%, surpassing state-of-the-art by 0.06\%. While these improvements seem minimal, on two datasets where the state-of-the-art is near perfect, ConstellationNet's improvements are still significant, as evidenced by its increase of over 3 times compared to the previous state-of-the-art's improvement and being comparable to the prior state of the art's improvement on CIFAR 10. 

ConstellationNet achieves a competitive accuracy of 91.10\% on the ImageNet dataset, not outperforming OmniVec but still beating most other state-of-the-art. This result isn't indicative of ConstellationNet's performance on the entire dataset due to having only trained on 100 thousand samples and testing on the next 20 thousand. While the results suggest that ConstellationNet performs strongly on simpler datasets and scales worse on complex and large-scale datasets, ConstellationNet's small size and training time indicate that its potential to improve is high. 

While ConstellationNet performs above state of the art on two datasets tested, its main benefit comes from its size, augmented predictions, and lower training time. On MNIST, ConstellationNet has 1.8 million parameters, similar to the Merging CNN, but only needs to train for around five epochs, whereas the Merging CNN trains for 300 epochs\cite{merging}. Additionally, on CIFAR 10, compared to the efficient adaptive ensemble, ConstellationNet's 5.8 million parameters and 20 epoch training time are almost half of its 11 million parameters and 5 times faster than its 100 epochs of training time\cite{ensemble}. Lastly, on Imagenet, despite using a large ConvNet\cite{convnet} with 198 million parameters, meaning that ConstellationNet totals 203 million parameters, ConstellationNet still performs better than transformer models like CoCa and Model Soups, both with over 2000 million parameters, while training for up to 100 epochs, over 9 times faster than OmniVec, which trains for 900 epochs in addition to pretraining for 2000 epochs\footnote{OmniVec size not listed due to not being state in paper.}\cite{coca, soup, omnivec}.

Overall, the results confirm the power and efficiency of ConstellationNet in image classification tasks, outperforming state-of-the-art with significantly less training time and parameters.

\subsection{Unsupervised Results}
\subsubsection{Baselines}
Before discussing ConstellationNet's results across clustering datasets, we first demonstrate the GNN's ability to improve clustering, demonstrated in \cref{baselines} for MNIST and CIFAR 10:

\begin{table}[h!]
\centering
\begin{tabular}{|l|cc|cc|}
\hline

{Model} & \multicolumn{2}{c|}{MNIST} & \multicolumn{2}{c|}{CIFAR-10} \\
                       & NMI   & ARI   & NMI   & ARI   \\
\hline
GCN                    & 0.676 & 0.573 & 0.127 & 0.078 \\
GIN                    & \textbf{0.771} & \textbf{0.712 }& 0.111 & 0.062 \\
GAT                    & 0.710 & 0.647 & 0.131 & 0.080 \\
SGAT                   & 0.695 & 0.606 & \textbf{0.137} & \textbf{0.086} \\
K-means                & 0.491 & 0.360 & 0.079 & 0.041 \\
Agglomerative          & 0.693 & 0.453 & 0.077 & 0.036 \\
\hline
\end{tabular}
\caption{Baseline Performance for Models on MNIST and CIFAR-10}
\label{baselines}
\end{table}

Across both datasets, when compared to the best-performing baselines, GNN models using the DMoN loss operator can perform significantly better than baselines, with top models improving NMI by 1.1x on MNIST and 1.7x on CIFAR-10 while improving ARI by 1.6x on MNIST and over 2x on CIFAR-10. Base GNN models and baselines both have somewhat good performances on MNIST due to the separated dataset and low dimensionality, but both GNN models and baselines suffer on CIFAR-10 due to the higher dimensionality of the dataset. Thus, while the GNN and DMoN operator perform better than baselines, a basic GNN still suffers similarly from the curse of dimensionality, showcasing the strength of the DMoN operator and the neighborhood aggregations of the GNN while presenting a limitation that ConstellationNet aims to resolve. 
\subsubsection{Ablation Study}

Table \ref{ablation} presents an ablation study exploring different configurations of ConstellationNet on MNIST and CIFAR-10, with both the base constellationNet and its use as a transform tested across both datasets.

\begin{table}[h]
    \centering
    \begin{tabular}{@{}lcccc@{}}
        \textbf{Configuration} & \multicolumn{2}{c}{\textbf{MNIST}} & \multicolumn{2}{c}{\textbf{CIFAR-10}} \\
                               & NMI & ARI & NMI & ARI \\
        \hline
        Kmeans                      & 0.49 & 0.36 & 0.08 & 0.04 \\
        \hspace{0.3cm} +UMAP        & 0.86 & 0.83 & 0.08 & 0.04 \\
        \hspace{0.3cm} +UMAP + GCN  & 0.91 & 0.92 & -    & -    \\
        \hspace{0.3cm} +DINO        & -    & -    & 0.80 & 0.60 \\
        \hspace{0.3cm} +DINO + UMAP & -    & -    & 0.86 & 0.81 \\
        \hspace{0.3cm} +DINO + UMAP + GCN  & -    & -    & 0.90 & 0.86 \\
        \hspace{0.9cm} \textit{(ConstellationNet)} \\
        GCN + DMoN                  & 0.68 & 0.57 & 0.13 & 0.08 \\
        \hspace{0.3cm} +UMAP        & 0.92 & 0.93 & -    & -    \\
        \hspace{0.3cm} +DINO        & -    & -    & 0.72 & 0.72 \\
        \hspace{0.3cm} +DINO + UMAP & -    & -    & 0.90 & 0.93 \\
        \hspace{0.9cm} \textit{(ConstellationNet)} \\
        \hline
    \end{tabular}
    \caption{ConstellationNet ablation studies across MNIST and CIFAR-10. Each component is tested using both the final GNN and Kmeans.}
    \label{ablation}
\end{table}

We observe that incorporating UMAP enhances performance in the case of an already separated dataset, confirming its use as a feature enhancer. On MNIST, where clusters are already somewhat distinct, UMAP offers a very concrete advantage, with the combination of UMAP and a GNN achieving an NMI of 0.92 and an ARI of 0.93, surpassing all other configurations and competing with state-of-the-art clustering. UMAP similarly improves the NMI and ARI of Kmeans on MNIST but doesn't affect its NMI or ARI on CIFAR-10, a dataset where data points are not distinctly grouped. Given that the UMAP functions by projecting manifolds down to a lower dimension, its poor performance on a dataset where no clear manifolds can be found is somewhat expected. 

In addition to UMAP, using DINO significantly enhances the performance of both the Kmeans and GNN on CIFAR-10. DINO isn't used on MNIST because it is dissimilar to the training set for DINO. The DINO operator's ability to embed data points and separate clusters significantly improves clustering performance and allows UMAP to further enhance features, as seen in the best NMI of 0.90 and an ARI of 0.93. 

Finally, using the final GNN as an embedding, ConstellationNet is shown to have concrete and significant improvements over a baseline like Kmeans. On MNIST, passing data through ConstellationNet before the Kmeans causes an improvement of around two times for both NMI and ARI and over 10 times for CIFAR-10, showcasing the power of ConstellationNet as a transform. 

\subsubsection{ConstellationNet}
To evaluate ConstellationNet's clustering performance, we test it against deep learning based image clustering state-of-the-art on CIFAR-10, CIFAR-100, and Oxford Flowers, as seen in \cref{table:performance_comparison}. 
\begin{table}[h]
\centering
\scalebox{0.85}{
\begin{tabular}{l|ccc|ccc|ccc}
\hline
Model & \multicolumn{3}{c|}{CIFAR-10} & \multicolumn{3}{c|}{CIFAR-100} & \multicolumn{3}{c}{Flowers} \\
& NMI       & ARI       & Accuracy     & NMI       & ARI       & Accuracy     & NMI       & ARI       & Accuracy     \\
\hline
KMeans  & 0.08      & 0.04      & 0.06         & 0.021     & 0.142     & 0.213        & 0.048     & 0.195     & 0.203        \\
ConstellationNet      & \textit{0.931}     & \textit{0.944  }   & \textbf{0.971}        & \textit{0.876 }    & 0.124     & \textbf{0.967   }     & 0.992     & 0.971     & \textbf{0.999} \\
TURTLE                 & \textbf{0.985} & \textbf{0.989} & \textit{0.969} & \textbf{0.915} & \textbf{0.832} & \textit{0.899}   & 0         & 0         &\textit{ 0.996 }       \\
TEMI CLIP ViT-L        & 0.926     & 0.932     & 0.946        & 0.799     & \textit{0.612 }    & 0.737        & -        & -         & -            \\
DPAC  & 0.87      & 0.866     & 0.934 & -        & -         & -     & -        & -         & -   \\
SPICE*    & -        & -         & -     & 0.583     & 0.422        & 0.584     & -         & -            & -            \\
\hline
\end{tabular}
}
\caption{Performance of Different Models on CIFAR-10, CIFAR-100, and Flowers Datasets. The best results are highlighted, and the second-best results are italicized.}
\label{table:performance_comparison}
\end{table}

ConstellationNet demonstrates leading performance on three datasets, outperforming most other state-of-the-art except unsupervised transfer, or TURTLE. On CIFAR-10 and CIFAR-100, ConstellationNet closely trails TURTLE while outperforming it in accuracy, and on the Flowers dataset, ConstellationNet surpasses all other available state-of-the-art, achieving a near-perfect accuracy of 0.999 compared to TURTLE's 0.996. Thus, ConstellationNet demonstrates its power amongst the current state of the art, being competitive across all metrics. 

However, the strength of ConstellationNet again lies in its smaller model size and training time. Due to TURTLE and TEMI being the closest-performing models, we only compare ConstellationNet to them when referring to runtime and parameters\cite{turtle,temi}. Both TURTLE and TEMI use a DINO v2 as a data embedding before their respective frameworks, but TURTLE notably uses a DINO v2 giant instead of a DINO v2 large, which results in an increase of around 500 million parameters compared to our framework. TURTLE also uses CLIP as another representation space, which can add another 400 million parameters to the model size. For the clustering model, ConstellationNet uses around 500 thousand trainable parameters for its GNN, while TEMI uses another vision transformer or DINO model, both several hundred million parameters, and TURTLE uses upwards of 1.9 million parameters\cite{temi,turtle}. Thus, the unsupervised ConstellationNet pipeline consists of around 500 million parameters, while TURTLE uses upwards of 1.5 billion parameters, and TEMI uses around 1 billion parameters. Additionally, while ConstellationNet trains for 100 epochs, TEMI trains for 200 epoch, and TURTLE trains for 6000 iterations at a batch size of 10000, which for CIFAR-10 and 100 equates to 1000 epochs. This means that ConstellationNet competes with state-of-the-art with more a twofold reduction in parameters and training, showcasing its robust power across datasets. 

Overall, these results confirm the power and efficiency of ConstellationNet in unsupervised image clustering tasks, being able to efficiently cluster both as an independent pipeline and as a transform for another clustering method with less memory and time used.

\section{Conclusion}\label{sec13}

In conclusion, this study has introduced ConstellationNet, a CNN-GNN framework that performs state-of-the-art clustering while enhancing baselines by over 10 times. Using both a convolutional neural network and a graph neural network, ConstellationNet addresses significant problems of dimensionality and local information problems that traditional and deep learning clustering methods face while using mini-batching to improve runtime and predictions. Through novel message passing and residual edge connection frameworks, ConstellationNet showcases its power as an end-to-end clustering pipeline and data embedding, outperforming state-of-the-art across several popular image datasets in accuracy, size, and training time. Due to its robust supervised and unsupervised performance, fast predictions due to smaller model size and minibatching, and data-specific knowledge, ConstellationNet holds many potential applications, able to be pretrained and quickly deployed in fields like epidemiology and medical imaging to solve arising problems. In addition to this use, ConstellationNet also holds promise as a replacement for other clustering methods in time-sensitive applications, such as search engines and anomaly detection. Future work in the area involves creating a better unsupervised loss for clustering, which can adapt to unbalanced datasets.

\bmhead{Acknowledgements}

We thank Julian Shun for providing datasets during the experimental process and high-level feedback during points of the research process.

\section*{Declarations}

\begin{itemize}
\item Funding: No Funding was given for the project.
\item Conflict of interest/Competing interests: The authors have no competing interests to declare that are relevant to the content of this article.
\item Ethics approval and consent to participate: NA
\item Consent for publication: All authors consent to the publication of this article. 
\item Data availability: All data used is taken from publically available datasets.
\item Code availability: All code for the project is located in the following github: https://github.com/hydrus3109/Nature-ConstellationNet
\item Author contribution: Junhong acted as the mentor for the project, and Aidan conducted experimentation. 
\end{itemize}

\noindent

\bibliography{References}

\end{document}